\documentclass[onecolumn,11pt]{IEEEtran}
\usepackage{times}

\usepackage[utf8]{inputenc} %
\usepackage[T1]{fontenc}    %
\usepackage{hyperref}       %
\usepackage{url}            %
\usepackage{booktabs}       %
\usepackage{amsfonts}       %
\usepackage{nicefrac}       %
\usepackage{microtype}      %
\usepackage[ruled]{algorithm2e}

\usepackage{xcolor}

\usepackage{amsmath}
\usepackage{subcaption}
\usepackage{enumitem}

\usepackage{mathtools}
\DeclareMathOperator*{\argmin}{argmin} %

\title{Neurally Augmented ALISTA}

\usepackage[]{graphicx}

\author{%
  Freya Behrens$^*$,
  Jonathan Sauder$^*$
  and
  Peter Jung\\
  {\small \{f.behrens,sauder\}@campus.tu-berlin.de
    and peter.jung@tu-berlin.de
  }
  \thanks{$\mbox{}^*$ First two authors have equal contribution}
}

\begin{document}

\maketitle

\begin{abstract}
  It is well-established that many iterative sparse reconstruction algorithms can be unrolled to yield a learnable neural network for improved empirical performance.
  A prime example is learned ISTA (LISTA) where weights, step sizes and thresholds are learned from training data.
  Recently, Analytic LISTA (ALISTA) has been introduced, combining the strong empirical performance of a fully learned approach like LISTA, while retaining theoretical guarantees of classical compressed sensing algorithms and significantly reducing the number of parameters to learn.
  However, these parameters are trained to work in expectation, often leading to suboptimal reconstruction of individual targets.  In this work we therefore introduce Neurally Augmented ALISTA, in which an LSTM network is used to compute step sizes and thresholds individually for each target vector during reconstruction.
  This adaptive approach is theoretically motivated by revisiting the recovery guarantees of ALISTA.
  We show that our approach further improves empirical performance in sparse reconstruction, in particular outperforming existing algorithms by an increasing margin as the compression ratio becomes more challenging.
\end{abstract}

\section{Introduction and Related Work}

Compressed sensing deals with the problem of recovering a sparse
vector from very few compressive linear observations, far less than
its ambient dimension. Fundamental works of Candes, Romberg, Tao and Donoho \cite{compressedsensing,Donoho2006a} show
that this can be achieved in a robust and stable manner with
computationally tractable algorithms given that the observation matrix
fulfills certain conditions, for an overview see \cite{foucart}.
Formally, consider the set of $s$-sparse vectors in $\mathbb{R}^N$,
i.e.
$\Sigma_s^N := \big\{ x \in \mathbb{R}^N \big| \|x\|_0 \leq s\big\}$
where the size of the support of $x$ is denoted by $\|x\|_0:=|\text{supp}(x)| = |\{i\,:\,x_i\neq 0\}|$. Furthermore, let $\Phi %
\in \mathbb{R}^{M \times N}$ be the
measurement matrix, with typically $M \ll N$. For a given noiseless
observation $y=\Phi x^*$ of an unknown but $s$-sparse
$x^* \in \Sigma_s^N$  we therefore wish to solve:
\begin{equation}\label{l0_reconstruction}\argmin_x {{\|x\|}_0} \text{\hspace{10px}s.t.\hspace{10px}} y = \Phi x\end{equation}
In \cite{compressedsensing} it has been shown, that under certain assumptions on $\Phi$,
the solution to the combinatorial problem in \eqref{l0_reconstruction} can be also obtained by a convex relaxation where one instead
minimizes the $\ell_1$--norm  
of $x$.
The Lagrangian formalism
yields then an unconstrained optimization problem
also known as LASSO \cite{lasso}, which penalizes
the $\ell_1$-norm via the hyperparameter $\lambda \in \mathbb{R}$:
\begin{equation}\label{l1_reconstruction}
\hat{x} = \argmin_x \frac{1}{2}\|y - \Phi x\|_2^2 + \lambda\|x\|_1
\end{equation}
A very popular approach for solving this problem is the iterative shrinkage thresholding algorithm (ISTA) \cite{ista}, in which a reconstruction $x^{(k)}$ is obtained after $k$ iterations from initial $x^{(0)}=0$ via the iteration:
\begin{equation}\label{ista_alg}
x^{(k + 1)}=\eta_{\lambda/L}\Big(x^{(k)} + \frac{1}{L}\Phi^T(y - \Phi x^{(k)})\Big)
\end{equation}
where $\eta_\theta$ is the soft thresholding function given by $\eta_\theta(x) = \text{sign}(x)\max(0, |x|-\theta)$ (applied coordinate-wise) and $L$ is the
Lipschitz constant (i.e. the largest eigenvalue) of $\Phi^T\Phi$. %
Famously, the computational graph of ISTA with $K$ iterations can be unrolled to yield Learned ISTA (LISTA) \cite{lista}, a $K$-layer neural network in which all parameters involved can be trained (each layer $k$ has an individual threshold parameter and individual or shared matrix weights) using backpropagation and gradient descent. LISTA achieves impressive empirical reconstruction performance for many sparse datasets but loses the theoretical guarantees of ISTA.
Bridging the gap between LISTA's strong reconstruction quality and the
theoretical guarantees for ISTA, ALISTA \cite{alista} was
introduced. ALISTA, introduces a matrix $W^T$, related to the measurement matrix $\Phi^T$ in \eqref{ista_alg}, which is computed by optimizing the generalized coherence:
\begin{equation}\label{mu_coh}
  \mu(W, \Phi) = \inf_{W \in \mathbb{R}^{M \times N}}\max_{i \neq j} W_{:,i}^T \Phi_{:,j} \text{\hspace{5pt}s.t.\hspace{3pt}} \forall i \in \{1, \dots , N\}: W_{:,i}^T \Phi_{:,i}=1
\end{equation}
Then, contrary to LISTA, all matrices are excluded from learning in order to retain desirable properties such as low coherence. 
For each layer of ALISTA, only a scalar step size parameter $\gamma^{(k)}$ and a scalar threshold $\theta^{(k)}$ is learned from the data, yielding the iteration:
\begin{equation}\label{alista}
x^{(k+1)} = \eta_{\theta^{(k)}}\Big(x^{(k)} - \gamma^{(k)} W ^T(\Phi x^{(k)} - y)\Big)
\end{equation}
As in LISTA, the parameters for ALISTA are learned end-to-end using backpropagation and stochastic gradient descent by empirically minimizing the reconstruction error:

\begin{equation}\label{mse}
\min_{\theta^{(1)}, \dots, \theta^{(K)}, \gamma^{(1)}, \dots , \gamma^{(K)}} \mathbb{E}_{x^*}\Big[ \|x^{(K)} - x^*\|_2^2\Big]
\end{equation}

The authors rigorously upper-bound the reconstruction error of ALISTA in the noiseless case and demonstrate strong empirical reconstruction quality even in the noisy case. The empirical performance similar to LISTA, the retained theoretical guarantees, and the reduction of number of parameters to train from either $O(KM^2 + NM)$ in vanilla LISTA or $O(MNK)$ in the variant of LISTA-CPSS \cite{listacpss} to just $O(K)$, %
make ALISTA an appealing algorithm to study and extend.

In \cite{learn_stepsizes}, instead of directly focusing on the reconstruction problem, where $\lambda$ is not known a priori, analytical conditions for optimal step sizes in ISTA are derived for LASSO, yielding Stepsize-ISTA. Stepsize-ISTA is a variant of LISTA in which the measurement matrices are exempt from training like in ALISTA, outperforming existing approaches to directly solving LASSO.

Thresholds that are adaptive to the current target vector have been
explored in ALISTA-AT \cite{other_adaptive}.
Following the majorization-minimization method, component-wise
thresholds are computed from previous iterations. In a particular
case this yields $\theta_i^{(k)}=1 / (1 + |x^{(k-1)}_i| / \epsilon)$
for some $\epsilon>0$, known as iterative reweighted
$\ell_1$-minimization. By unrolling this algorithm, the authors demonstrate superior recovery over
ALISTA for a specific setting of $M,N$ and $s$.

In a related approach \cite{gatedISTA} identify undershooting, meaning that reconstructed components are smaller than target components, as a shortcoming of LISTA and propose Gated-LISTA to address these issues. The authors introduce gain and overshoot gates to LISTA, which can amplify the reconstruction after each iteration before and after thresholding, yielding an architecture resembling GRU cells \cite{gru}. The authors demonstrate better sparse reconstruction than previous LISTA-variants and also show that adding their proposed gates to ALISTA, named AGLISTA, it is possible to improve its performance in the same setting of $M,N$ and $s$ as ALISTA-AT.

In this paper, motivated by essential proof steps of ALISTA's recovery guarantee, we propose an alternative method for
adaptively choosing thresholds and step sizes during reconstruction.
Our method directly extends
ALISTA by using a recurrent neural network to predict thresholds and step sizes
depending on an estimate of the $\ell_1$-error between the
reconstruction and the unknown target vector after each iteration. We refer to our method as Neurally Augmented ALISTA (NA-ALISTA), as the method falls into the general framework of neural augmentation of unrolled algorithms \cite{neuralaugmention, surveyUnrolling, unrolled}. 
The rest of the paper is structured as follows: we provide theoretical motivation for NA-ALISTA in Section~\ref{motivation_section}, before describing our method in detail in Section~\ref{naalista_section}. In Section~\ref{experiments_section}, we demonstrate experimentally that NA-ALISTA achieves state-of-the-art performance in all evaluated settings.
To summarize, our main contributions are:

\begin{enumerate}[leftmargin=*]
    \item We introduce Neurally Augmented ALISTA (NA-ALISTA), an algorithm which learns to adaptively compute thresholds and step-sizes for individual target vectors during recovery. The number of parameters added does not scale with the problem size.
    \item We provide theoretical motivation inspired by guarantees for sparse reconstruction which show that NA-ALISTA can achieve arrive tighter error bounds depending on the target $x^*$.
    \item We find that NA-ALISTA empirically outperforms ALISTA and other state-of-the-art algorithms in all evaluated settings and that the gains increase with decreasing $M/N$.  
\end{enumerate}

\section{Theoretical Motivation}\label{motivation_section}

The thresholds $\theta^{(k)}$ in \eqref{alista} play an important role in the analysis of ALISTA. While the authors of \cite{alista} prove that $\theta^{(k)}$ must be larger than a certain value in order to guarantee no false positives in the support of the reconstruction $x^{(k)}$, the thresholds $\theta^{(k)}$ also appear as an additive term in the reconstruction error upper bound.
Thus, to guarantee good reconstruction $\theta^{(k)}$ should be just slightly %
larger than the value it must surpass in order to both minimize the error and verify the assumption. In this section, we repeat key insights from ALISTA and motivate the choice of adaptive thresholds - the key improvement in our proposed NA-ALISTA. More specifically, we repeat the conditions under which ALISTA guarantees no false positives and highlight an intermediate step in the error bound from \cite{alista}, which tightens when the thresholds can adapt to specific instances of $x^*$.

\paragraph{Assumption }(adapted from Assumption 1 from \cite{alista})\label{assumption}\\
\textit{Let $x^* \in \Sigma_s^N$ be a fixed $s$--sparse target vector. Let $W$ be such that it attains the infimum of the generalized coherence with $\Phi$ (as in \eqref{mu_coh}) and denote this generalized coherence as $\tilde{\mu} = \mu(W, \Phi)$. Let $s < (1 + 1/\tilde{\mu}) / 2$. Let $\gamma^{(1)}, \dots , \gamma^{(K)}$ be any sequence of scalars taking values in $(0, \frac{2}{2\tilde{\mu}s - \tilde{\mu} + 1})$ and $\theta^{(1)}, \dots , \theta^{(K)}$ with}: 

\begin{equation}\label{ineq}\theta^{(k)}\geq \gamma^{(k)} \tilde{\mu} \| x^{(k)} - x^* \|_1\end{equation}

Because in ALISTA, the thresholds $\gamma^{(1)}, \dots , \gamma^{(K)}$ and stepsizes  $\theta^{(1)}, \dots, \theta^{(K)}$  are optimized in expectation
over the training data, the inequality in (\ref{ineq}) holds only in the
general case if the thresholds are larger than the worst case
$\ell_1$-error committed by the algorithm over all training vectors $x^*$ i.e.:

\begin{equation}\label{ineq_sup}\theta^{(k)}\geq \gamma^{(k)} \tilde{\mu} \sup_{x^*} \| x^{(k)} - {x^*} \|_1 \end{equation}

This is needed in order to fulfill the Assumption. Under these conditions it is guaranteed that no false positives are in the support of the reconstruction:

\paragraph{No false positives} (Lemma 1 from \cite{alista})\\
\textit{Under the settings of the Assumption, it holds that:} 
\begin{equation}\text{supp}(x^{(k)}) \subseteq \text{supp}(x^*)\end{equation}
\vskip 0.1in

However, the threshold $\theta^{(k)}$ also reappears in the error upper bound. Here we employ an intermediate step of the error upper bound from \cite{alista}:

\paragraph{Reconstruction error upper bound}(Theorem 1 from \cite{alista})\\
\textit{Under the settings of the Assumption, it holds that:}
\begin{equation}\label{errorbound}\|x^{(k +1)} - x^*\|_2 \leq \|x^{(k +1)} - x^*\|_1 \leq \tilde{\mu} \gamma^{(k)}(s -1) \|x^{(k)} - x^*\|_1 + \theta^{(k)}s + |1 - \gamma^{(k)}| \|x^{(k)} - x^*\|_1\end{equation}

 Where the first inequality holds for all real vectors and the second
 inequality is derived in detail in Appendix A of \cite{alista}.
  It is therefore desirable that
$\theta^{(k)}$ is as small as possible, but such that it still
satisfies \eqref{ineq}. This means that ALISTA has to learn
thresholds at least proportional to the largest possible committed
$\ell_1$-error over all possible $x^*$ in order to guarantee good
reconstruction, for which it is in turn penalized in the error bound.%

However, the thresholds that make the error bound tighter vary
depending on the $x^*$ that is to be recovered. In fact, if an
algorithm would have access to $\|x^{(k)} - x^*\|_1$ and were allowed to choose thresholds adaptively, depending on this quantity, the more relaxed inequality (\ref{ineq}) could be employed directly, without taking the supremum. An algorithm which approximates such thresholds, resulting in a tighter error bound, is the aim of this paper.

\newpage
\section{Neurally Augmented ALISTA}\label{naalista_section}
\begin{figure}[b!]
\begin{center}
 \captionsetup[subfigure]{width=0.99\linewidth}
\begin{subfigure}{.245\linewidth}
\centerline{\includegraphics[width=0.99\linewidth]{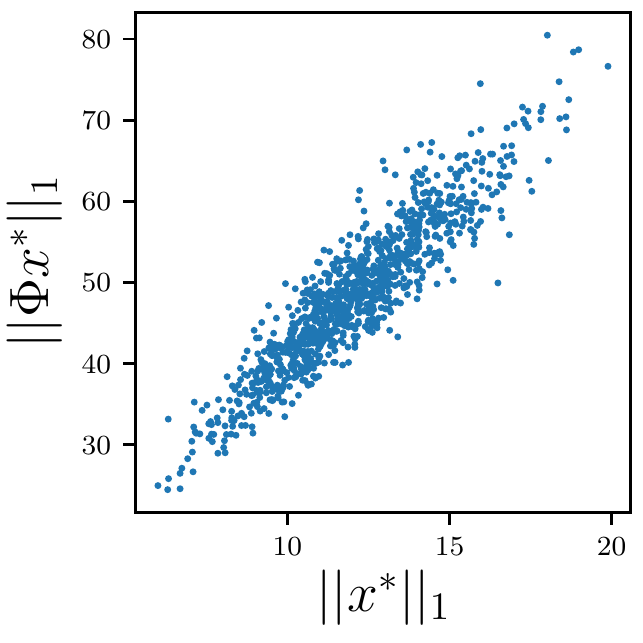}}
\subcaption{$r, \|x^*\|_0=15$}
\end{subfigure}
\begin{subfigure}{.245\linewidth}
\centerline{\includegraphics[width=0.99\linewidth]{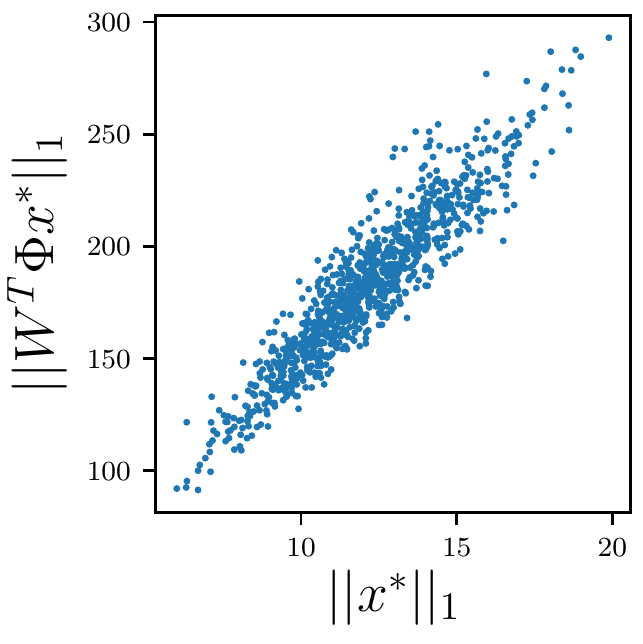}}
\subcaption{$u, \|x^*\|_0=15$}
\end{subfigure}
\begin{subfigure}{.245\linewidth}
\centerline{\includegraphics[width=0.99\linewidth]{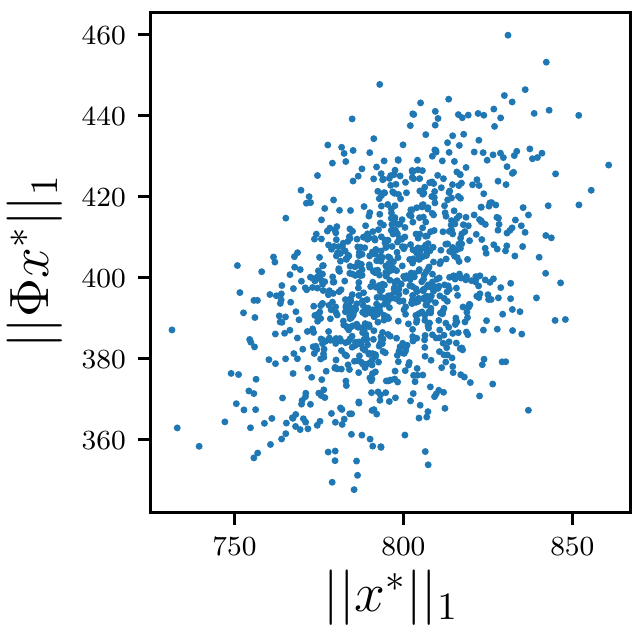}}
\subcaption{$r, \|x^*\|_0=1000$}
\end{subfigure}
\begin{subfigure}{.245\linewidth}
\centerline{\includegraphics[width=0.99\linewidth]{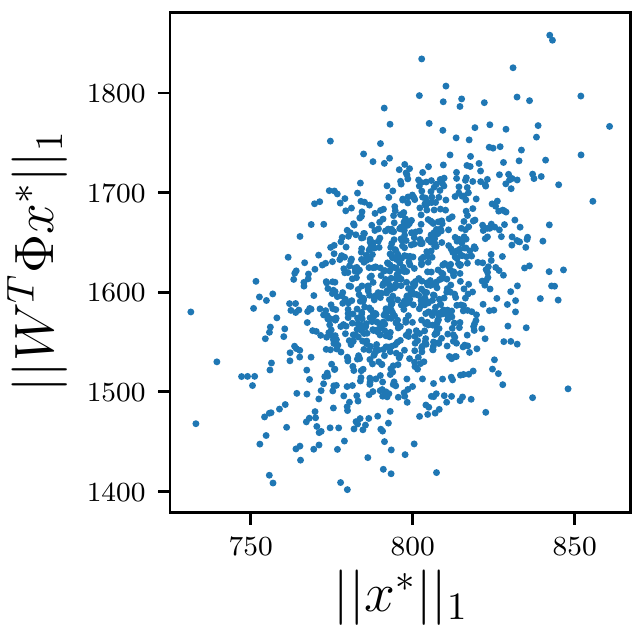}}
\subcaption{$u, \|x^*\|_0=1000$}
\end{subfigure}
\end{center}
\caption{Correlation between $\|x^*\|_1$ and $r=\|\Phi x^*\|_1$ and $u=\|W^T \Phi x^*\|_1$ for sparse vectors with $\|x^*\|_0=15$ (a) and (b) and non-sparse vectors $\|x^*\|_0=N$ (b) and (c). Nonzero components of $x^*$ are drawn i.i.d. from $\mathcal{N}(0, 1)$ with $N=1000$. One can see that for sparse $x^*$, $r$ and $u$ are correlated with $\|x^*\|_1$, whereas there is no obvious correlation for non-sparse vectors.} %
\label{correlation_l1}
\end{figure}

In order to tighten the error upper bound in 
\eqref{errorbound}, we introduce Neurally Augmented ALISTA
(NA-ALISTA), in which we adaptively predict thresholds
$\theta^{(k,x^*)}$ depending on the current estimate for the
$\ell_1$-error between $x^{(k)}$ and the unknown $x^*$. As
can be observed from \eqref{ineq}, such
$\theta^{(k,x^*)}$ must be proportional to $\|x^{(k)} -
x^*\|_1$. 

In theory, this true $\ell_1$-error could be recovered exactly. This is because there are no false
positives in $x^{(k)}$, making it $s$-sparse and for a $\tilde{\mu} < 1 / (2s - 1)$ the column-normalized
$W^T \Phi$ is restricted-invertible for any $2s$-sparse input \cite{foucart} [Corollary 5.4, p.113]. However, it is infeasible to solve such an inverse problem at every iteration $k$. Furthermore, in practice the sparsity is often much larger than what is admissible via the coherence bound. For example, in the experiments of \cite{lista, alista, gatedISTA, other_adaptive}, a sparsity of $50$ is used with $M$=250, $N$=500. This sparsity already exceeds a maximum admitted sparsity of 11 derived from the minimum theoretical coherence of 0.0447 by the Welch Bound \cite{welch}, implying that such an exact recovery is not possible in practice anyways.

NA-ALISTA is thus largely concerned with learning for each
iteration $k$ a good approximation of $\|x^{(k)} - x^*\|_1$. For this, consider the $\ell_1$-norms of the residual:
 \begin{equation}\label{residual}
     r^{(k)}:=\| \Phi x^{(k)} - y \|_1 = \| \Phi (x^{(k)} - x^*) \|_1
 \end{equation}
 and the iterative update quantity in \eqref{alista}:
 \begin{equation}\label{quantity}
     u^{(k)}:=\| W ^T(\Phi x^{(k)} - y) \|_1 = \| (W^T\Phi) (x^{(k)} - x^*) \|_1
 \end{equation}

\begin{figure}[t!]
\begin{center}
 \captionsetup[subfigure]{width=0.99\linewidth}
\begin{subfigure}{.245\linewidth}
\centerline{\includegraphics[width=0.99\linewidth]{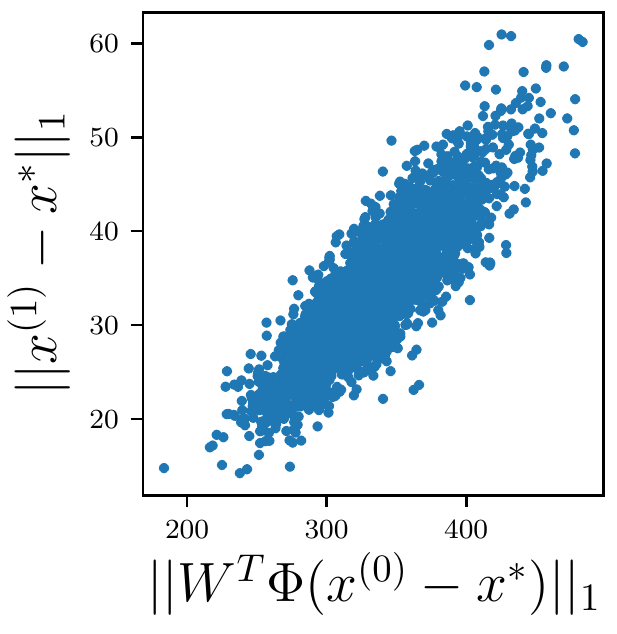}}
\subcaption{}
\end{subfigure}
\begin{subfigure}{.245\linewidth}
\centerline{\includegraphics[width=0.99\linewidth]{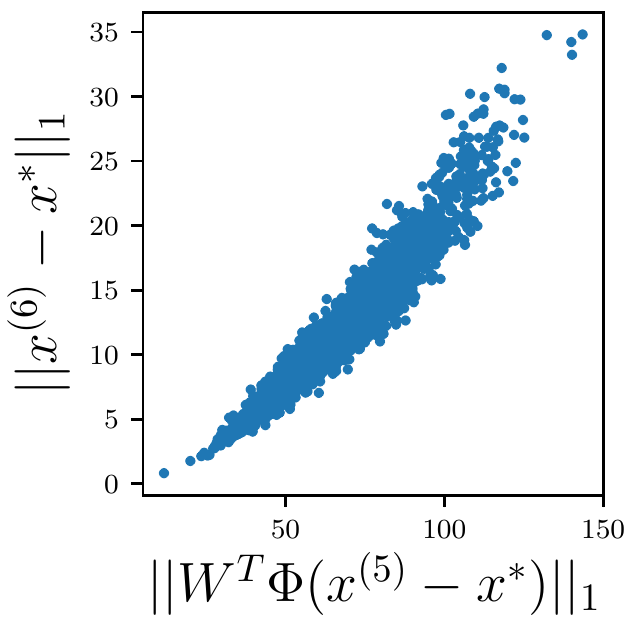}}
\subcaption{}
\end{subfigure}
\begin{subfigure}{.245\linewidth}
\centerline{\includegraphics[width=0.99\linewidth]{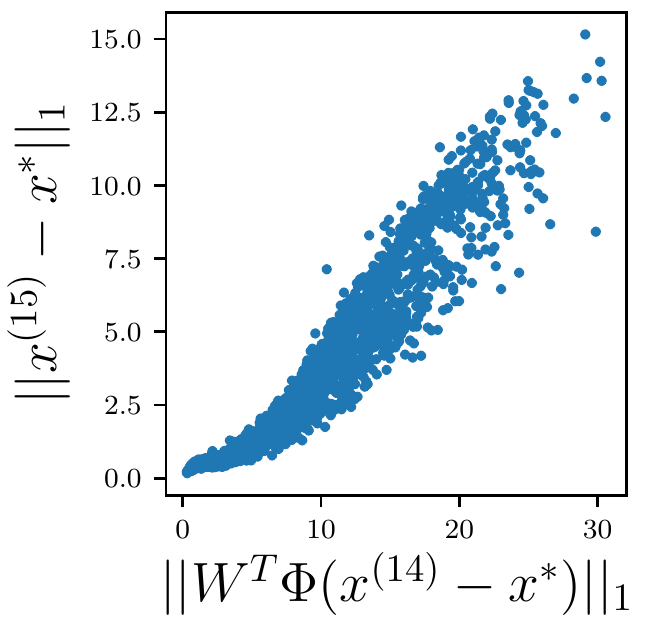}}
\subcaption{}
\end{subfigure}
\begin{subfigure}{.245\linewidth}
\centerline{\includegraphics[width=0.99\linewidth]{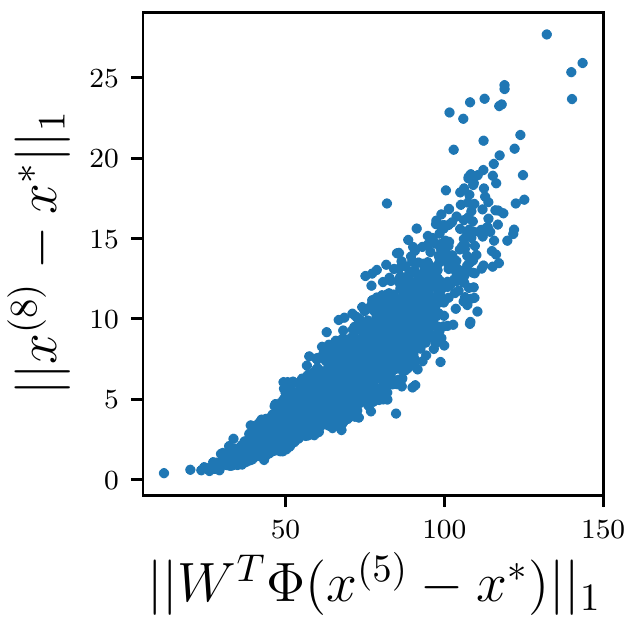}}
\subcaption{}
\end{subfigure}
\end{center}
\vskip -0.1in
\caption{Correlation between $u^{(i)}$ and $\|x^{(j)} - x^*\|_1$ in a trained instance of NA-ALISTA for $(i,j) = (0, 1)$ (a), $(5, 6)$ (b), $(14, 15)$ (c), $(5, 8)$ (d). There is a clear correlation, that is even preserved across multiple iterations (d), suggesting the use of a recurrent neural network to predict $\theta^{(k,x^*)}$. Training was performed with the settings described in Section \ref{experiments_section}, with $N=1000$, $H=128$ and $K=16$.}
\label{correlation_layers}
\end{figure}

Both are known to the algorithm even though $x^*$ is unknown. That $r^{(k)}$ and $u^{(k)}$ are useful quantities for approximating the true $\ell_1$-error stems from the fact that $W^T \Phi$ has low mutual coherence, thus being a restricted identity for sparse vectors. This is visualized in Figure \ref{correlation_l1}. 
Other useful quantities to approximate the true $\ell_1$-error are given by $\|x^{(0)} -x^*\|_1, \dots,\|x^{(k-1)} -x^*\|_1$. This is highlighted by Figure \ref{correlation_layers}
and suggests the use of a recurrent neural network in NA-ALISTA. We
therefore propose to use an LSTM \cite{lstm} which has two input
neurons, receiving $u^{(k)}$ and $r^{(k)}$ at each iteration $k$. This is used to update the internal state and produce the outputs $\theta^{(k, x^*)}$ and $\gamma^{(k, x^*)}$, which are used to compute the next iteration, producing the update rule:

\begin{equation}\label{naalista}
x^{(k+1)} = \eta_{\theta^{(k, x^*)}}\Big(x^{(k)} - \gamma^{(k, x^*)} W ^T(\Phi x^{(k)} - y)\Big)
\end{equation}

A computational expression for NA-ALISTA is given in Algorithm \ref{algorithm}.
Note that the introduction of LSTM-cells in NA-ALISTA does not significantly increase the required computing power in practice. In fact, in Section \ref{experiments_section}, we show that small LSTM-cells suffice for best empirical performance, independently of the problem size. Let $H$ be the size of the hidden layer of the LSTM-cells, then the computation for a single forward computation of the cell takes $O(H^2)$ computations. As a regular iteration of ALISTA takes $O(MN)$ operations and computing the $\ell_1$-norm of the update quantity $W^T (\Phi x^{(k)} - y)$ takes an additional $O(N)$ operations, an iteration of NA-ALISTA requires $O(MN + N + H^2)$ operations. For example, when $M=250, N=2000, H=64$ as in one of the experimental settings in Figure \ref{hidden_layer}, then $H^2/MN=4096/500000=0.008192$, showing that the added computation is negligible in practice. %

\begin{algorithm}[t!]
\SetAlgoLined
 \textbf{Learnable Parameters:} initial cell state $c_0 \in \mathbb{R}^H$, initial hidden state $h_0 \in \mathbb{R}^H$, \\
 cell state to output matrix $ U \in  \mathbb{R}^{2 \times H}$ and parameters of LSTM cell.\\
 \textbf{Input:} $y$\\
 $x\leftarrow0$; $h\leftarrow h_0$; $c\leftarrow c_0$\\
 \For{$\{1,  \dots , K\}$}{
  $r \leftarrow \|\Phi x - y\|_1$\\
  $u \leftarrow \|W^T (\Phi x - y) \|_1$\\
  $c, h \leftarrow \texttt{LSTM}(c, h, [r, u])$\\
  $\theta, \gamma \leftarrow \texttt{Softsign}(U c)$\\
  $x \leftarrow\eta_{\theta}\Big(x - \gamma W^T(\Phi x - y)\Big)$
 }
\textbf{Return} $x$;
 \caption{Neurally Augmented ALISTA}
 \label{algorithm}
\end{algorithm}

\section{Experiments}\label{experiments_section}
In this section, we evaluate NA-ALISTA in a sparse reconstruction task and compare it against ALISTA \cite{alista}, ALISTA-AT \cite{other_adaptive}, AGLISTA \cite{gatedISTA}, as well as the classical ISTA \cite{ista} and FISTA \cite{fista}. To emphasize a fair and reproducible comparison between the models, the code for all experiments listed is available on GitHub
\footnote{\url{https://github.com/feeds/na-alista}}

\subsection{Experimental Setup}
Following the same experimental setup as \cite{alista, gatedISTA,
  listacpss, other_adaptive}, the support of $x^*\in \mathbb{R}^N$ is
determined via i.i.d. Bernoulli random variables with parameter
$S/N$, leading to an expected sparsity of $S$. The non-zero components of $x^*$ are then sampled according to
$\mathcal{N}(0, 1)$. The entries of $\Phi$ are also sampled from
$\mathcal{N}(0, 1)$, before each column is normalized to unit
$\ell_2$-norm. 
$W$ is then computed by minimizing the generalized coherence in \eqref{mu_coh} between $W$ and $\Phi$ via the Frobenius-Norm approximation using projected
gradient descent. 
This procedure is identical to \cite{alista, gatedISTA, other_adaptive}.
The Adam optimizer \cite{adam} is used to minimize the $\ell_2$-error
from \eqref{mse} for all algorithms. A test set of 10000
samples is fixed before training and recovery performance is measured
with the normalized mean squared error (NMSE):
$$\text{NMSE} = 10 \log_{10}\Big(\frac{\mathbb{E}_{x^*}[\|x^{(K)}- x^*\|^2]}{\mathbb{E}_{x^*}[\|x^*\|^2]}\Big)$$
A support selection trick was introduced in \cite{listacpss} to speed
up convergence and stabilize training and has been subsequently used
extensively in variants LISTA and ALISTA \cite{alista, other_adaptive,
  gatedISTA}.
When support selection is used, a hyperparameter $p=(p^{(1)}, \dots , p^{(K)})$ is set such that for
each layer, a certain percentage of the largest absolute values are
exempt from thresholding, i.e.:
$$\eta_{(\theta,p^{(k)})}(x)_i = \begin{cases}x_i, &\text{if}\hspace{10pt} |x_i| \geq \lfloor p^{(k)}/N \rfloor\text{-largest value of } |x|\\
\text{sign}(x_i)\max(0, |x_i|-\theta) &\text{else}\\
\end{cases}$$
For a fair comparison, we employ support selection in all learned models compared in this paper similarly to the literature \cite{alista, listacpss, gatedISTA, other_adaptive}.
Our AGLISTA implementation follows the description in the paper \cite{gatedISTA}: we use exponential gain gates and inverse-proportional-based overshoot gains. The $\lambda$ parameter in ISTA and FISTA was tuned by hand, we found that $\lambda=0.4$ led to the best performance in our tasks. NA-ALISTA by default
uses both $r^{(k)}$ and $u^{(k)}$ as inputs to the LSTM in iteration $k$.

When not otherwise indicated we use the following settings for
experiments and algorithms: $M$=250, $N$=1000, $S$=50, $K$=16,
$H$=128, and $y=\Phi x^*+z$ with additive white Gaussian noise $z$ with a signal to noise ratio
SNR$:=\mathbb{E}(\|\Phi x^*\|_2^2)/\mathbb{E}(\|z\|_2^2)=40$dB.
We train all algorithms for 400 epochs, with each epoch containing
50,000 sparse vectors with a batch size of 512.

\begin{figure}[b!]
\begin{center}
 \captionsetup[subfigure]{width=0.99\linewidth}
\begin{subfigure}{.49\linewidth}
\centerline{\includegraphics[width=0.99\linewidth]{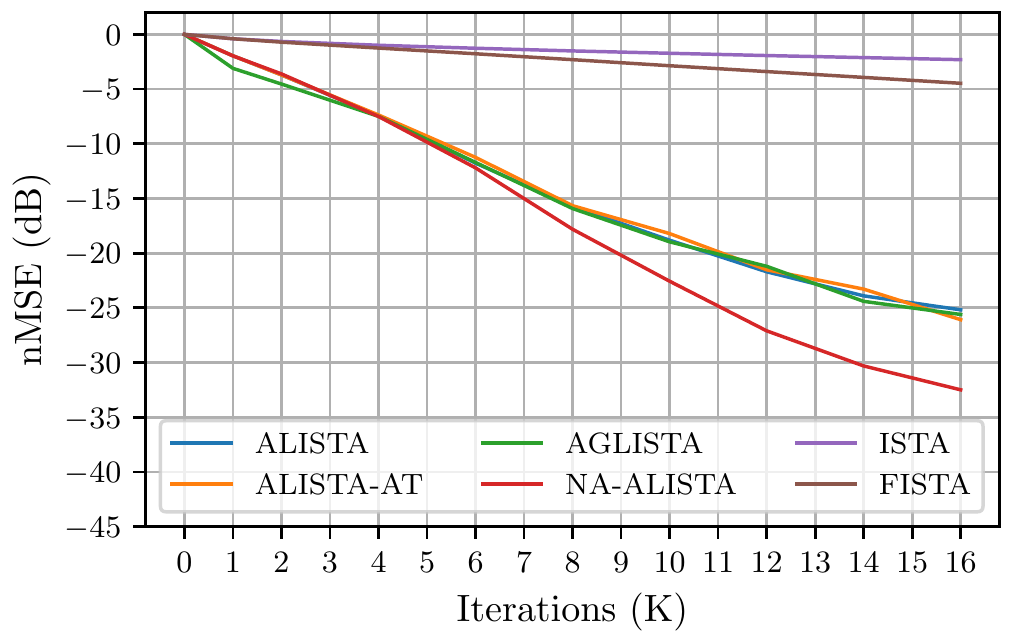}}
\subcaption{$N$=2000, SNR=$40$dB}
\label{varyk2000:2000_40}
\end{subfigure}
\begin{subfigure}{.49\linewidth}
\centerline{\includegraphics[width=0.99\linewidth]{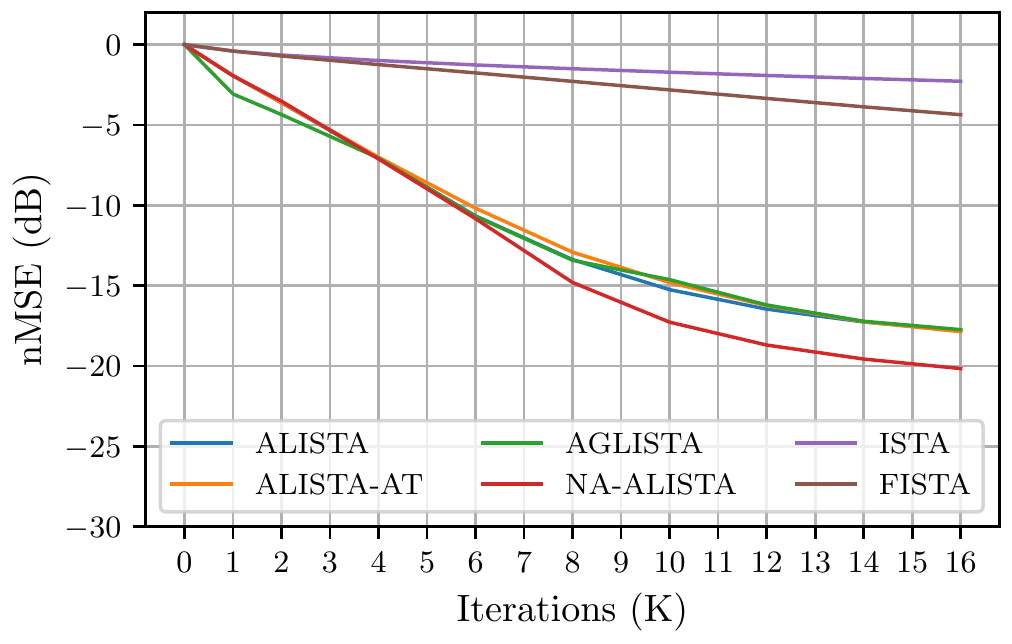}}
\subcaption{$N$=2000, SNR=$20$dB}
\label{varyk2000:2000_20}
\end{subfigure}
\end{center}
\vskip -0.1in
\caption{ The reconstruction error for ALISTA, AGLISTA, ALISTA-AT and NA-ALISTA over the number of iterations $K$ for SNR=$40$dB (\ref{varyk2000:2000_40}) and SNR=$20$dB (\ref{varyk2000:2000_20}). NA-ALISTA outperforms all competitors. Results for settings with smaller $N$ can be found in Appendix A.}
\label{varyk2000}
\end{figure}

\subsection{Comparison with Competitors}

As an established experimental setting to compare the performance of  of ISTA-based methods the compressed sensing, previous work  \cite{alista,other_adaptive,gatedISTA} has focused on a compression level of $ M/N=0.5$ with sparsity $S$=50 following \cite{listacpss}. 
However, practical applications in communication and imaging favor even lower compression rates
like $10\dots20\%$, which is why we extend our analysis to more challenging rates. To achieve different compression rates we keep the sparsity $S$ and measurements $M$ constant while increasing $N$. 

\begin{figure}[t]
\vskip -0.15in
\begin{center}
 \captionsetup[subfigure]{width=0.99\linewidth}
 \begin{subfigure}{.49\linewidth}
\centerline{\includegraphics[width=\linewidth]{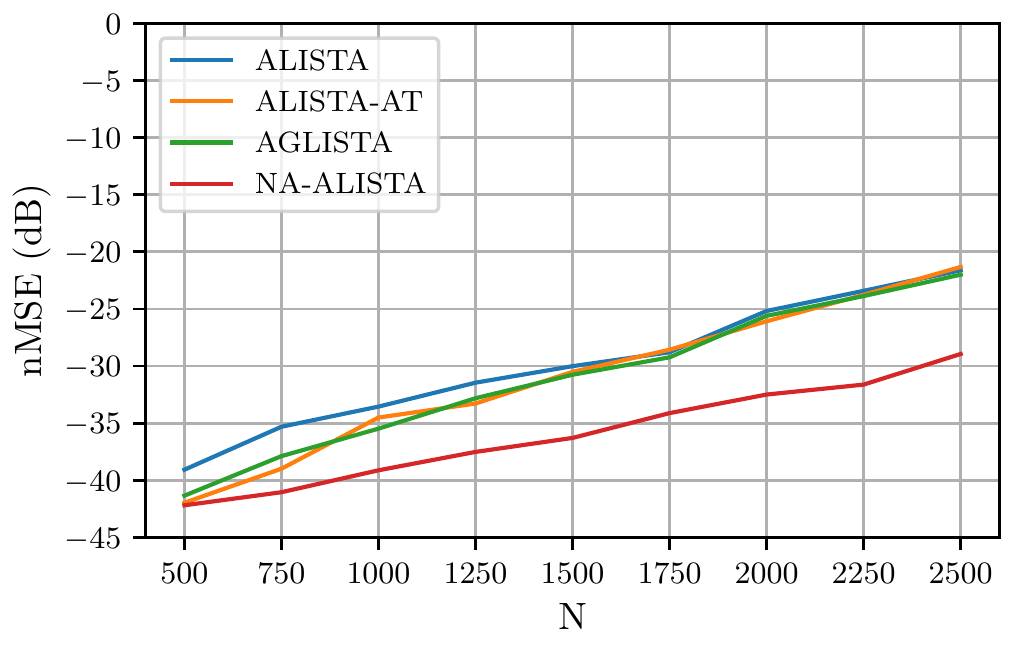}}\vskip -0.05in
\subcaption{SNR=$40$dB\label{varyn40}}

\end{subfigure}
\begin{subfigure}{.49\linewidth}
\centerline{\includegraphics[width=\linewidth]{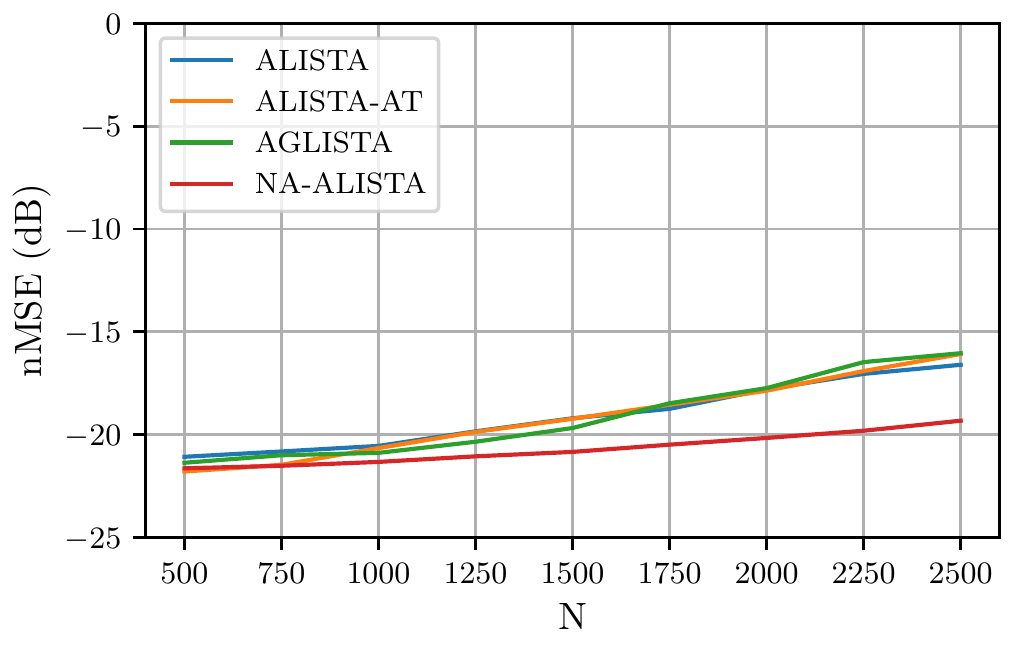}}\vskip -0.05in

\subcaption{SNR=$20$dB\label{varyn20}}

\end{subfigure}
\end{center}
\vskip -0.14in
\caption{Reconstruction error over  different compression ratios. For a constant expected sparsity of $S$=50 and $M$=250 measurements and $K$=16 iterations, the input size $N$ varies. Both under a SNR of $40$dB and $20$dB NA-ALISTA increases its reconstruction margin to competitors as  $N$ increases and the compression ratio becomes more challenging.}
\label{varyn}
\end{figure}

As shown in Figure~\ref{varyk2000}, we first fix $N=2000$ and observe the reconstruction error for a varying amount of iterations. 
In Figure~\ref{varyn} we then decrease the compression ratio while keeping the sparsity constant. We observe that NA-ALISTA outperforms state-of-the-art adaptive methods in all evaluated scenarios. 
Whereas for the more established setting from the literature of $N$=500, the improvement of NA-ALISTA is small, this margin increases as the compression ratio becomes more challenging. 
In Figure~\ref{varyn40} the reconstruction error achieved by ALISTA-AT and AGLISTA deteriorates to the performance of ALISTA, while our NA-ALISTA can sustain its advantage over ALISTA even for compression rates up to $0.1$ when $N=2500$. This suggests that our method is interesting to a wider range of practical applications.

\begin{figure}
    \centering
    \begin{minipage}[t]{0.48\textwidth}
        \begin{center}
\centerline{\includegraphics[width=\linewidth]{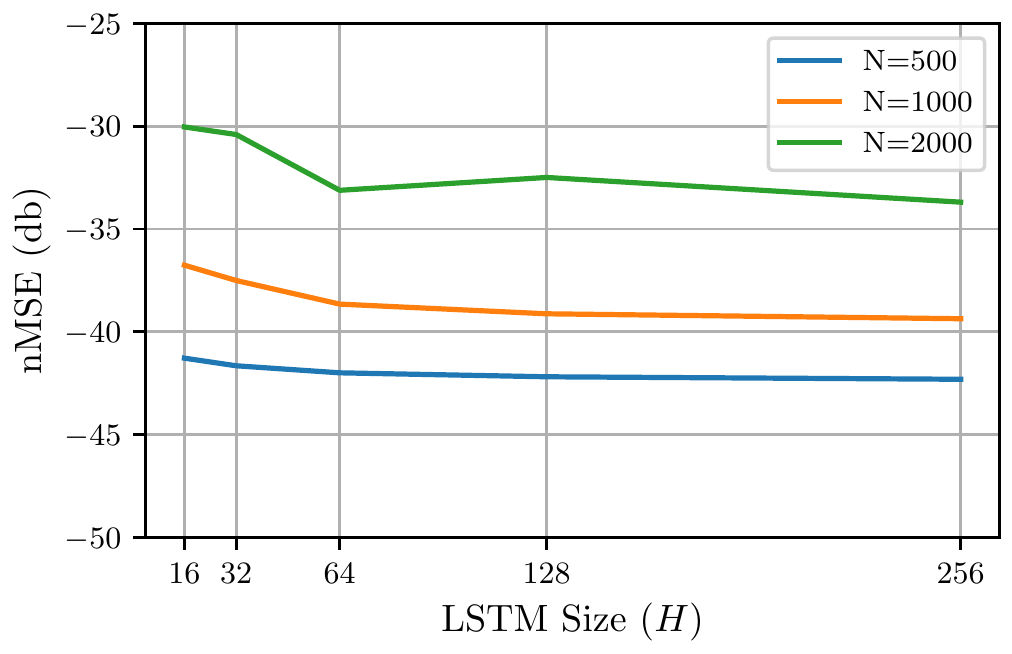}}
\vskip -0.22in
\end{center}
\caption{Reconstruction error for varying settings of the LSTM size in NA-ALISTA. Larger $N$ profit more from larger $H$, but in all settings an exponential increase of the LSTM size only yields a marginal improvement once $H$=64 is surpassed.}
\label{hidden_layer}
    \end{minipage}\hfill
    \begin{minipage}[t]{0.48\textwidth}
        \begin{center}
\centerline{\includegraphics[width=0.95\linewidth]{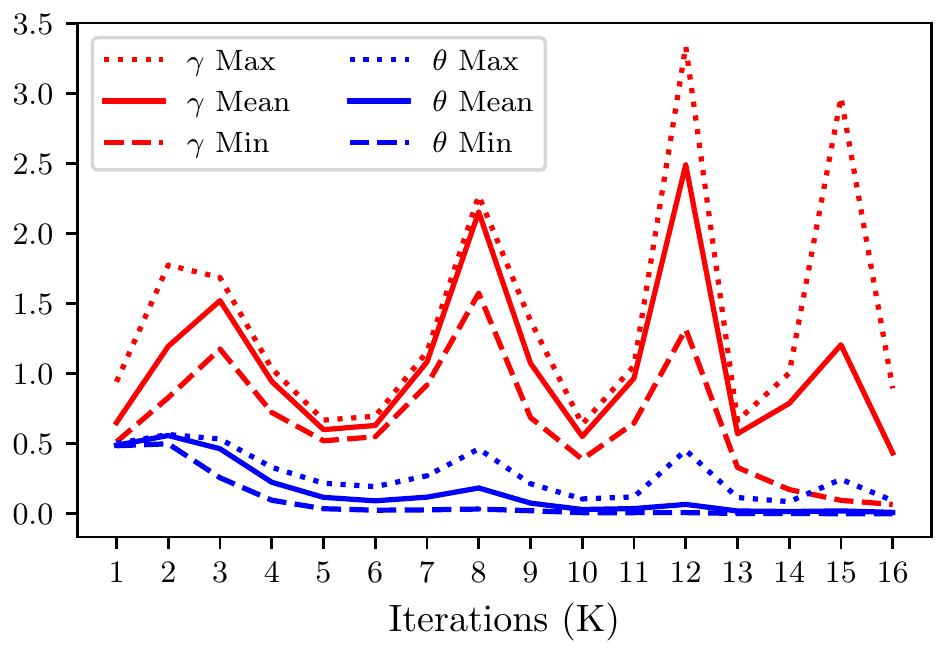}}
\vskip -0.22in
\end{center}
\caption{Predicted step sizes and thresholds from a trained instance of NA-ALISTA ($N=1000, M=250, S=50, K=16$), highlighting the adpativity of NA-ALISTA. Inference to obtain these values is performed on the test set.}
\label{gammas_and_thetas_are_adaptive}
    \end{minipage}
    \vskip -0.1in
\end{figure}

\begin{table}[b!]
\vskip -0.15in
    \centering
    \begin{tabular}{lrrr}
\toprule
Model  &       $N$=500 &       $N$=1000 &       $N$=2000 \\
\midrule
NA-ALISTA $r^{(k)}$  & -42.00 & -39.15 & \textbf{-32.50} \\NA-ALISTA $\{r^{(k)},u^{(k)}\}$ & \textbf{-42.18} & -39.12 & -32.49 \\NA-ALISTA $u^{(k)}$ & -42.03 & \textbf{-39.24} & -29.36 \\
\bottomrule
\end{tabular}
\caption{Reconstruction error in dB for NA-ALISTA with different
  inputs $r^{(k)}$ and/or $u^{(k)}$ to the LSTM (see
  \eqref{residual} and \eqref{quantity}) %
  with $K$=16, SNR=40. It does not matter which quantities we use to
  estimate the $\ell_1$-error, since all perform equally well as input
  to the LSTM.}
    \label{tab:my_label}
\vskip -0.1in

\end{table}

 To verify that the added computation, determined by the size $H$ of the LSTM, is negligible in practice, we test different settings of $H$. In Figure \ref{hidden_layer} we show that an exponential increase in hidden neurons yields only a small error reduction for different $N$, suggesting that the size $H$=128 is a sufficient default value for several settings of $N$. This implies that neural augmentation only marginally affects the runtime. We tested NA-ALISTA using different inputs $r^{(k)}, u^{(k)}$ for the LSTM in Table \ref{tab:my_label} and conclude that all approximations perform similarly and a single approximation of the $\ell_1$-error is sufficient. However, we observe a slight increase in convergence speed and training stability when using both inputs.

\begin{figure}
\vskip -0.15in
    \centering
        \begin{minipage}[t]{0.48\textwidth}
        \begin{center}
\centerline{\includegraphics[width=0.92\linewidth]{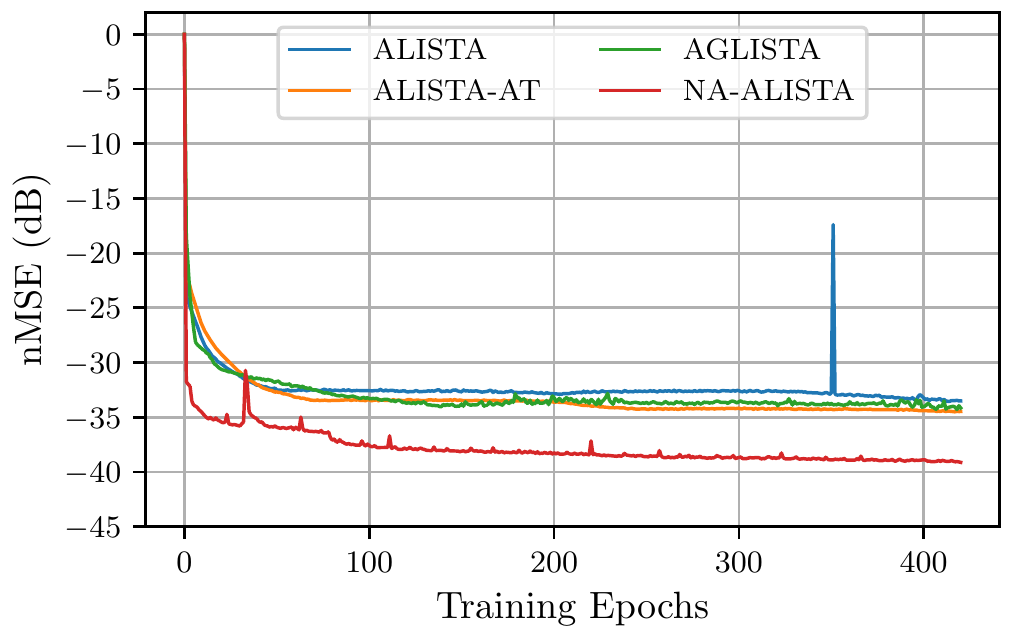}}
\vskip -0.2in
\end{center}
\caption{Training curves for $N$=1000, $K$=16, SNR=40 from the learned algorithms we compare in this paper, showing that NA-ALISTA outperforms the competitors after only a few epochs of training. Each epoch consists of 50,000 randomly drawn sparse vectors.}
\label{learning_curve}
    \end{minipage}\hfill
    \begin{minipage}[t]{0.48\textwidth}
        \begin{center}
\centerline{\includegraphics[width=\linewidth]{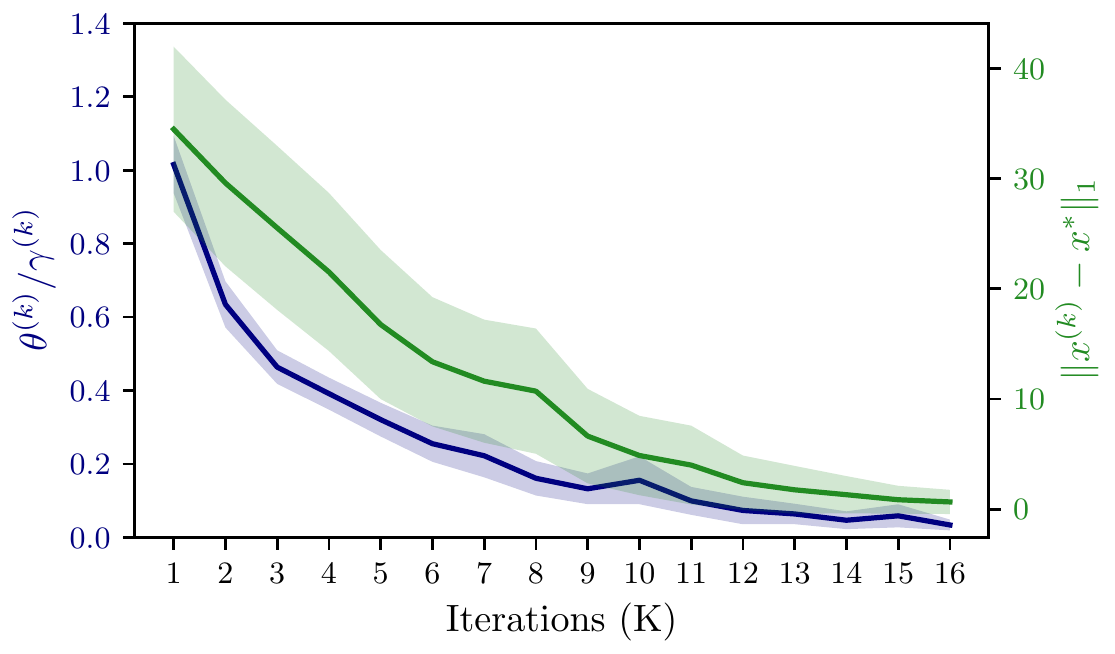}}
\vskip -0.2in
\end{center}
\caption{Comparison of the ratio $\theta^{(k)}/\gamma^{(k)}$ with the true $\ell_1$-error $||x^*-x^{(k)}||_1$ at each iteration for NA-ALISTA. We report the mean for a batch of randomly drawn test data $\{x^*\}$ along with the standard deviation for each quantity. Together these terms behave as desired, see Eq. \eqref{ineq} and its discussion.}%
\label{gammas_and_thetas_fullfill_the_assumptions}
    \end{minipage}
\vskip -0.1in
    \end{figure}

We also evaluate whether the increased empirical performance of NA-ALISTA is truly due to its adaptivity or simply due to its architecture, since the LSTM architecture could in principle enable a more stable optimization of the desired parameters due to more stable gradients. This would imply that when run on a test set, the learned step sizes would not vary depending on the input. Figure~\ref{gammas_and_thetas_are_adaptive} shows that this is not the case, since step sizes and thresholds vary within a margin on a test set~of~10,000 randomly sampled inputs. Also, the decreasing threshold $\theta^{(k)}$ corresponds to ``warm start'' behavior for ISTA to first go through a thresholding phase and then through a fitting phase where the threshold becomes essentially zero, see exemplary \cite{Loris_2009}.
An additional strength of NA-ALISTA is that it is fast and stable to train, outperforming competitors after only a few epochs, as shown in Figure~\ref{learning_curve}.

As an empirical verification of Assumption 1 in \eqref{ineq} we need to check for every $x^*$, whether the ratio $\theta^{(k,x^*)}/\gamma^{(k,x^*)}$ is proportional to the $\ell_1$-error $||x^*-x^{(k)}||_1$.
Since it is infeasible to check the assumption for the infinite set of sparse vectors $\Sigma_s^N$, we empirically verify \eqref{ineq} for a sample of inputs from the training distribution. In Figure \ref{gammas_and_thetas_fullfill_the_assumptions} the means of both values are proportional to each other for such a test sample, suggesting that the reconstruction bound \eqref{errorbound}  holds for NA-ALISTA.

\section{Conclusion and Future Work}
In this paper, we propose Neurally Augmented ALISTA (NA-ALISTA), an extension of ALISTA in which the step sizes and thresholds are predicted adaptively to the target vector by a neural network. Besides a theoretical motivation for NA-ALISTA, we experimentally demonstrate that it is able to outperform state-of-the-art algorithms such as ALISTA \cite{alista}, AGLISTA \cite{gatedISTA}, and ALISTA-AT \cite{other_adaptive} in sparse reconstruction in a variety of experimental settings. In particular, NA-ALISTA outperforms the existing algorithms by a wide margin in settings with a large compression. 

While in this paper we restrict ourselves to the classical compressed sensing setting, in which $s$-sparse vectors are reconstructed, neural augmentation provides a more flexible framework for incorporating additional knowledge into classical algorithms. Therefore, an interesting line of future work is to explore how neural augmentation can incorporate notions of structured sparsity or other constraints into sparse reconstruction. There is a plethora of signal processing algorithms, going much beyond variants of ISTA, proximal gradient methods, and even beyond sparse reconstruction in general, which lend itself to an interpretation of a neural network when unfolded \cite{surveyUnrolling}. Identifying algorithms which could benefit from neural augmentation in the way that ALISTA does is left as future work.

\newpage
\bibliography{references}
\newpage
\appendix
\section{Supplementary Experiments}
\begin{figure}[h]
\begin{center}
 \captionsetup[subfigure]{width=0.99\linewidth}
 \begin{subfigure}{.49\linewidth}
\centerline{\includegraphics[width=0.99\linewidth]{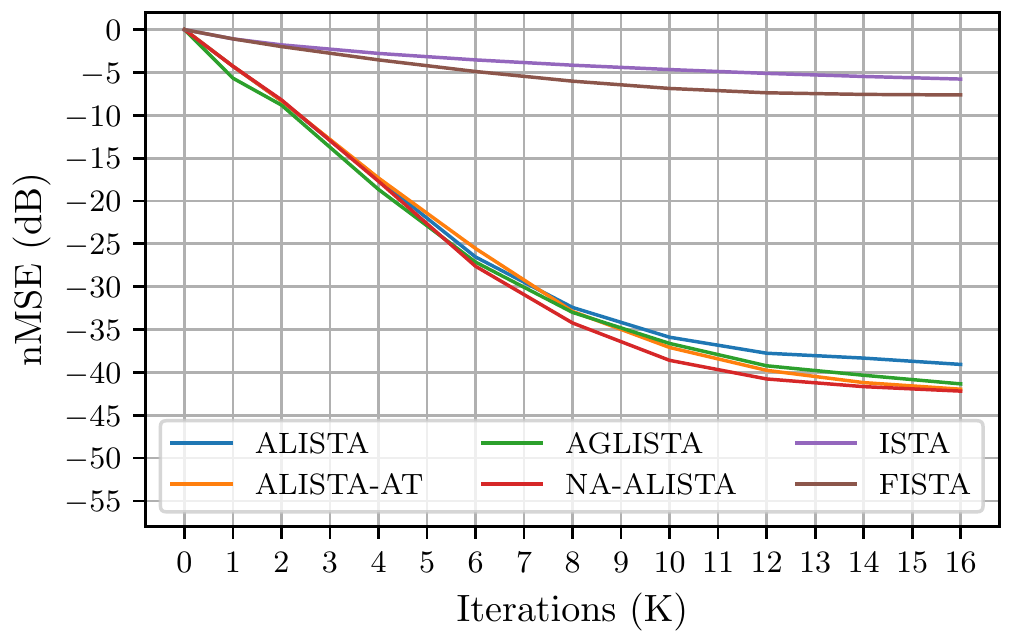}}
\subcaption{$N$=500, SNR=$40$}
\label{varyk:500_40}
\end{subfigure}
\begin{subfigure}{.49\linewidth}
\centerline{\includegraphics[width=0.99\linewidth]{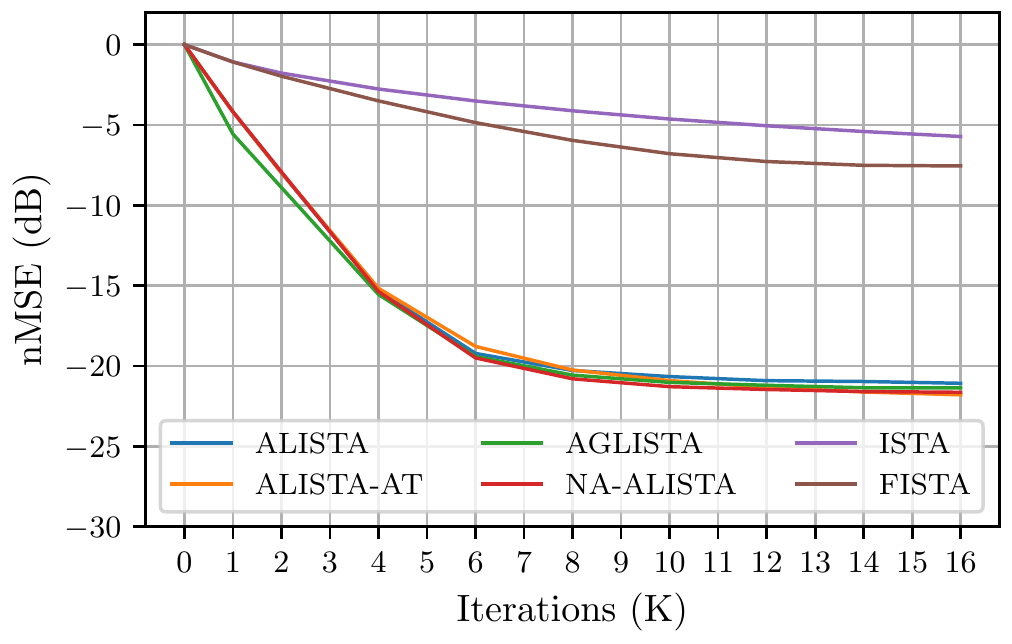}}
\subcaption{$N$=500, SNR=$20$}
\label{varyk:500_20}
\end{subfigure}
\begin{subfigure}{.49\linewidth}
\centerline{\includegraphics[width=0.99\linewidth]{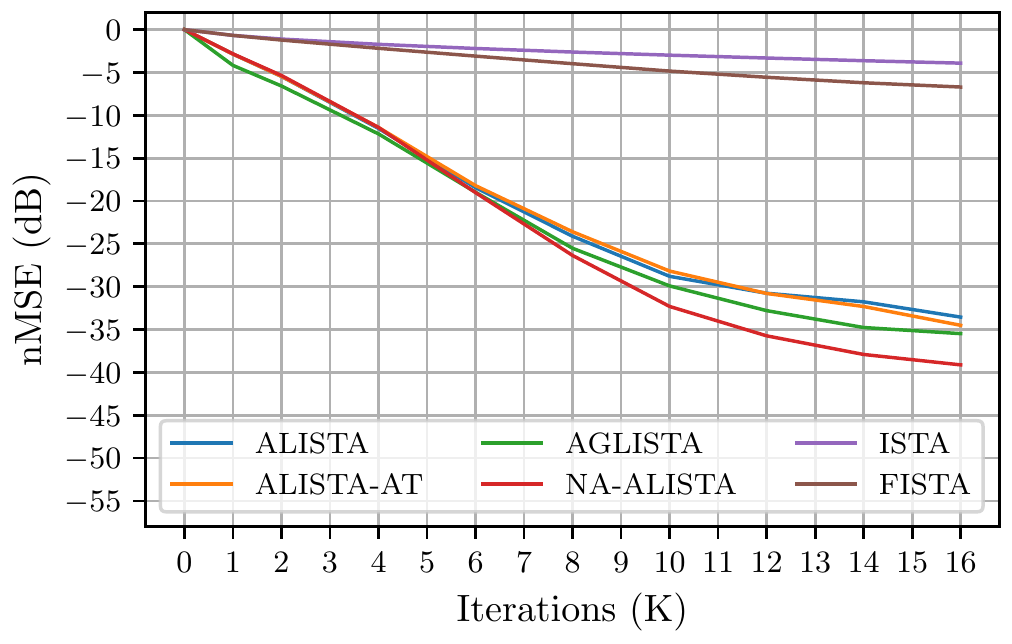}}
\subcaption{$N$=1000, SNR=$40$}
\label{varyk:1000_40}
\end{subfigure}
\begin{subfigure}{.49\linewidth}
\centerline{\includegraphics[width=0.99\linewidth]{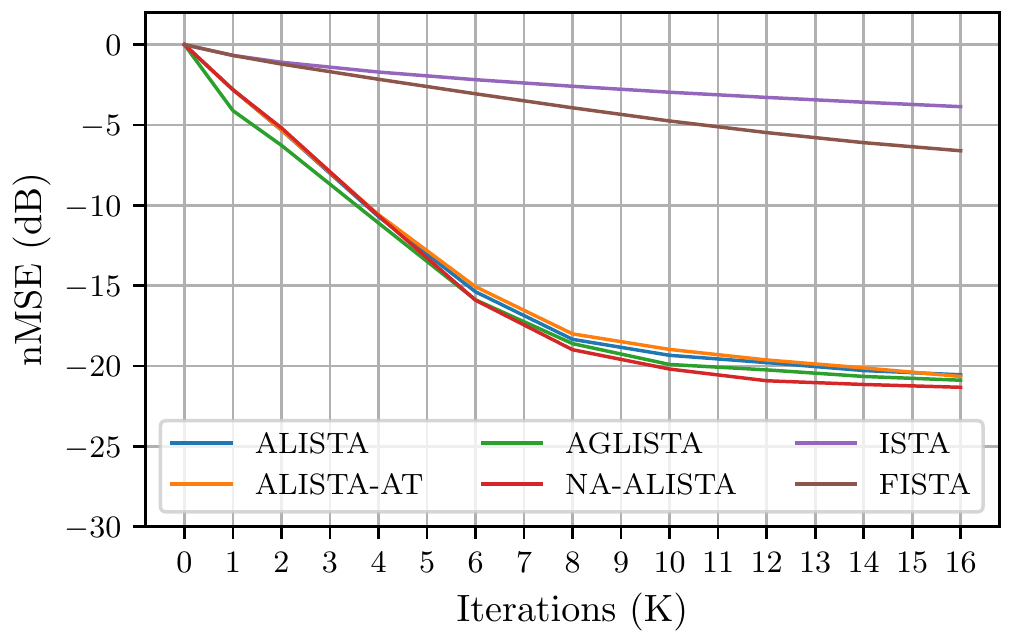}}
\subcaption{$N$=1000, SNR=$20$}
\label{varyk:1000_20}
\end{subfigure}
\begin{subfigure}{.49\linewidth}
\centerline{\includegraphics[width=0.99\linewidth]{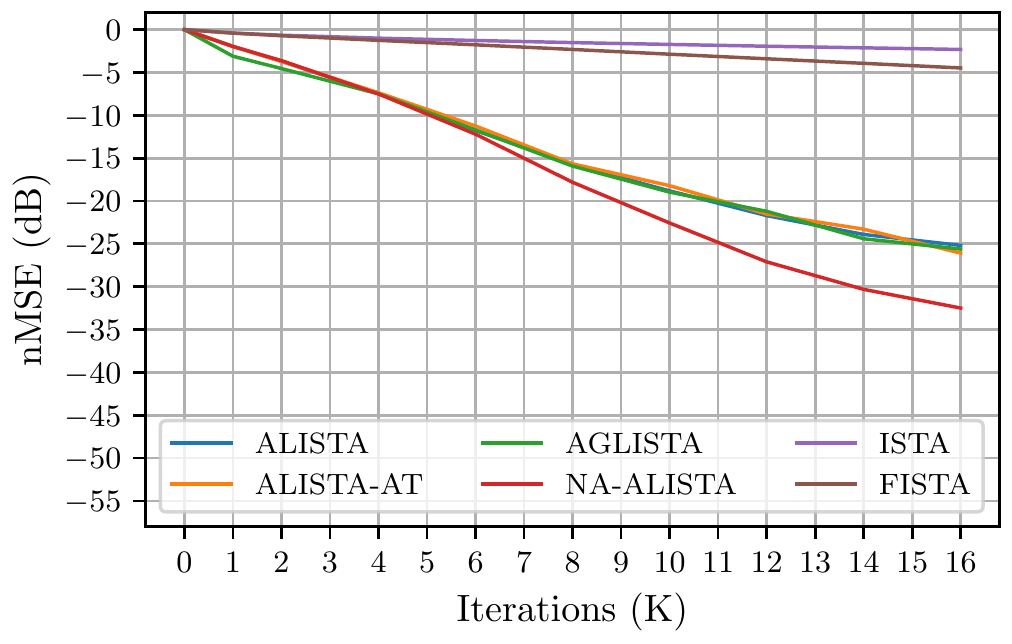}}
\subcaption{$N$=2000, SNR=$40$}
\label{varyk:2000_40}
\end{subfigure}
\begin{subfigure}{.49\linewidth}
\centerline{\includegraphics[width=0.99\linewidth]{res/varyk/varyk__n=2000_noise=GaussianNoise20.pdf}}
\subcaption{$N$=2000, SNR=$20$}
\label{varyk:2000_20}
\end{subfigure}
\end{center}
\vskip -0.1in
\caption{The reconstruction error for ALISTA, ALISTA-AT and NA-ALISTA over the number of iterations run for different noise and $N$ settings. In \ref{varyk:500_40}, for the standard setting in the literature with $N$= 500 and a noise level of $40$dB NA-ALISTA performs on par with competitors after 16 iterations. For an increased $N$=1000 under the same noise level in \ref{varyk:1000_40}, our algorithm outperforms the other methods clearly. For a noise level of 20dB all algorithms perform similarly for $N=$500 and $N$=1000 and NA-ALISTA outperforms the others at $N=2000$.}
\label{varyk}
\end{figure}

\end{document}